# POS Tagging Using Relaxation Labelling

Lluís Padró

Departament de Llenguatges i Sistemes Informàtics
Universitat Politècnica de Catalunya
Pau Gargallo, 5. Barcelona. 08071 Spain
e-mail : padro@lsi.upc.es

**Abstract**

Relaxation labelling is an optimization technique used in many fields to solve constraint satisfaction problems. The algorithm finds a combination of values for a set of variables such that satisfies -to the maximum possible degree- a set of given constraints. This paper describes some experiments performed applying it to POS tagging, and the results obtained. It also ponders the possibility of applying it to word sense disambiguation.

**Areas:** Natural Language Processing, POS tagging.

**Keywords:** POS tagging, HMM taggers, Relaxation labelling, Word Sense Selection.

**Category:** paper

# 1  Introduction and Motivation

Relaxation is a well-known technique used to solve consistent labelling problems. Actually, relaxation is a family of energy-function-minimizing algorithms closely related to Boltzmann machines, gradient step, and Hopfield nets.

A consistent labelling problem consists of, given a set of variables, assigning to each variable a value compatible with the values of the other ones, according to a set of compatibility constraints.

Relaxation has been traditionally used in computer vision: corner and edge recognition, line and image smoothing [Lloyd 83, Richards et al. 81], but many problems can be stated as a labelling problem: the travelling salesman problem, n-queens, or any other combinatorial problem. [Aarts & Korst 87]. The relaxation labelling algorithm is explained in section 2.

In this paper we will try to make a first insight into applying relaxation labelling to natural language processing. The main idea of the work is that NLP problems such as POS tagging or WSD can be stated as constraint satisfaction problems, thus, they could be addressed with the usual techniques of that field, such as relaxation labelling.

POS tagging consists of assigning to each word of a sentence a part-of-speech tag, that indicates the function of that word in that specific context. The current tendency is to model the sequence of tags in a sentence as a Hidden Markov Model, and try to compute the most probable tag sequence given the word sequence. To do this, one needs to estimate the model probabilities, such as the probability that a certain word appears with a certain tag, or the probability that a tag is followed by another. This estimation is usually done by computing frequencies on hand tagged corpora, like in [Garside et al. 87, Cutting et al. 92]. Iterative re-estimation of frequencies to improve model accuracy is also used in [Briscoe et al. 94, Elworthy 93, Cutting et al. 92]. Results are really good, giving about $95\% - 97\%$ of correctly tagged words. Some authors try to improve the results by using a set of context constraints which are applied to the results of the probabilistic tagger, and correct its most common errors [Brill 92, Moreno-Torres 94].

It seems reasonable to consider POS tagging or word sense disambiguation as combinatorial problems in which we have a set of variables (words in a sentence) a set of possible values for each one (POS tags or senses), and a set of constraints for these values. All we need to be able to use relaxation labelling is to find what these constraints are.

We might also consider the possibility of combining both problems in only one: words are variables, and each of them has two tags, a POS one and a sense one, (or a unique tag being a combination of both), then we would be able to express constraints between the two types of tags (for instance, that a verb with a sense of "movement" is highly compatible with a preposition. That would help to disambiguate the POS of a word that could be a preposition or a noun and was found after such a verb).

The application of relaxation labelling to POS tagging is exposed in section 3. The application to WSD is briefly discussed in section 5. The combined problem will be addressed in further work.

## 2 The Relaxation Labelling Algorithm

Relaxation operations have been long used in engineering fields to solve systems of equations [Southwell 40], but it got its biggest success when the extension to symbolic domain -Relaxation labelling- was applied to constraint propagation field, specially in low-level vision problems. [Waltz 75, Rosenfeld et al. 76]

Relaxation labelling is a generic name for a family of iterative algorithms which perform function optimization, based on local information. This relates them closely to neural nets and gradient step.

Let $V = \{v_1, v_2, \ldots, v_n\}$ be a set of variables

Let $t_i = \{t_1^i, t_2^i, \ldots, t_{m_i}^i\}$ be the set of possible values for variable $v_i$ (where $m_i$ is the number of values that $v_i$ can take).

Let $C$ be a set of constraints between the values of the variables. Each constraint is a "compatibility value" for a combination of variable values. For instance, $[(v_1, t_2^1)(v_3, t_1^3)] = 0.5$ is a constraint stating that the combination of variable $v_1$ having value $t_2^1$, and variable $v_3$ having value $t_1^3$ has a compatibility degree of 0.5. Constraints can be of any order, so we can define the compatibility degree for combinations of any number of variables (obviously we can have combinations of at most $n$ variables).

The aim of the algorithm is to find a weighted labelling such that "global consistency" is maximized. A weighted labelling is a weight assignment for each possible value of each variable: $P = (p^1, p^2, \ldots, p^n)$ where each $p^i$ is a vector containing a weight for each possible value of $v_i$, that is $p^i = (p_1^i, p_2^i, \ldots, p_{m_i}^i)$

The relaxation algorithm consists of:

- start in a random labelling $P_0$

- for each variable, compute the "support" that each value receives from the current weights for the values of the other variables (i.e. see how compatible is the current weighting with the current weightings of the other variables, given the set of constraints).

- Update the weight of each variable value according to the support obtained by each of them (that is, increase weight for values with high support, and decrease weight for those with low support).

- iterate the process until a convergence criterion is met.

The support computing and value changing must be performed in parallel, to avoid that changing a value of a variable would affect the support computation of the others.

We could summarize this algorithm saying that at each time step, a variable changes its value depending on how compatible is that value with the values of the other variables at that time step.

The effect of the algorithm is to maximize the weighted support value for each variable with respect to the constraint set and the values of the other variables. This is the above mentioned "global consistency".

Note that this concept makes the algorithm robust, since the problem of having mutually incompatible constraints (so one cannot find a combination of values which satisfies all the constraints) is solved because relaxation doesn't (necessarily) find an exclusive combination of values, that is, an unique value for each variable, but a weight for each possible value such that consistency is maximized (the constraints are satisfied to the maximum possible degree).

Being $p_j^i$ the weight for value $j$ in variable $v_i$ and $S_{ij}$ the support value for the same combination, relaxation maximizes, for all variables $v_i$
$\sum_j p_j^i \times S_{ij}$

The algorithm requires a way to compute which is the support for a variable value given the others and the constraints. This is called the "support function" and it is the heart of the algorithm, since it is what will be maximized.

Several support functions are used in the literature, depending on the problem addressed, to define the support received by tag $j$ of word $i$ ($S_{ij}$).

One of them is:
$$S_{ij} = \sum_{r \in R_{ij}} C_r \times p_{k_1}^{r_1}(m) \times \ldots \times p_{k_d}^{r_d}(m) \qquad (1.1)$$
where
$C_r$ is the compatibility coefficient for the constraint $r$
$R_{ij}$ is the set of constraints on tag $j$ for word $i$, that is:
$R_{ij} = \{r \mid r = [(v_{r_1}, t_{k_1}^{r_1}), \ldots, (v_i, t_j^i), \ldots, (v_{r_d}, t_{k_d}^{r_d})]\}$
$p_{k_l}^{r_l}(m)$ is the weight assigned to value $t_{k_l}^{r_l}$ for variable $v_{r_l}$ at time $m$.

Another possibility is:
$$S_{ij} = \prod_{d=1}^{D} \prod_{V \in V^d} \sum_{r \in R_{ij}^V} C_r \times p_{k_1}^{r_1}(m) \times \ldots \times p_{k_d}^{r_d}(m) \qquad (1.2)$$
where:
$D$ is the maximum degree for constraints (understanding a constraint degree as the number of variables involved in the constraint).

$V^d$ are all possible subsets of $d$ variables.
$R_{ij}^V$ is the set of $d$-degree constraints on tag $i$ for word $j$ in which the involved variables are those of $V$.

Yet another possibility:

$$S_{ij} = \prod_{d=1}^{D} \prod_{V \in V^d} \max_{r \in R_{ij}^V} \{C_r \times p_{k_1}^{r_1}(m) \times \ldots \times p_{k_d}^{r_d}(m)\} \quad (1.3)$$

The algorithm also needs to compute which is the new weight for a variable value, and this computation must be done in such a way that it can be proven to meet a certain convergence criterion, at least under appropriate conditions[1] [Zucker et al. 78, Zucker et al. 81, Hummel & Zucker 83].

This is called the "updating function" and it is used to compute and normalize the new weight values for each possible tag.

Several formulas have been proposed [Rosenfeld et al. 76], and some of them have been proven to be approximations of a gradient step algorithm.

The updating formulas must increase the weight associated with values with a higher support, and decrease those of values with lower support. This is achieved by multiplying the current weight of a value by a factor depending on the support received by that value. Normalization is performed in order that the weights for all the values of a variable add up to one.

Usual updating functions are:

This formula increases the weight for a value when $S_{ij}$ is positive and decreases it when $S_{ij}$ is negative. Values for $S_{ij}$ must be in $[-1, 1]$.

$$p_j^i(m+1) = \frac{p_j^i(m) \times (1+S_{ij})}{\sum_{k=1}^{k_i} p_k^i(m) \times (1+S_{ik})} \quad (2.1)$$

This formula increases the weight when $S_{ij} > 1$ and decreases it when $S_{ij} < 1$. Values for $S_{ij}$ must be nonnegative.

$$p_j^i(m+1) = \frac{p_j^i(m) \times S_{ij}}{\sum_{k=1}^{k_i} p_k^i(m) \times S_{ik}} \quad (2.2)$$

The following formula is also used:

$$p_j^i(m+1) = \frac{e^{S_{ij}/T}}{\sum_{k=1}^{k_i} e^{S_{ij}/T}} \quad (2.3)$$

---

[1] Convergence has been proven under certain conditions, but in a complex application such as POS tagging we will find cases where it is not necessarily achieved. Alternative stopping criterions will require further attention.

where $T$ is a temperature parameter which decreases at each time step. The labelling is non-ambiguous in this case (weights are only 0 or 1) and what we compute is the probability that a variable changes its value. When $T$ is high, changes occur randomly. As $T$ decreases, support values get more influence.

If variables take only one value at each time step (that is, one value has weight 1, and the others 0) and updating function (2.3) is used, the procedure is called stochastic relaxation (which is equivalent to simulated annealing), while if value weights are not discrete, and the updating function is (2.1) or (2.2) we talk about continuous deterministic relaxation.[2]

Relaxation requires the values of the support function to be nonnegative when using (2.2) and in $[-1, 1]$ when using (2.1). Negative values for (2.2) are avoided just not using that updating formula with negative compatibility values, since it makes no sense to do so. Normalizing in $[-1, 1]$ is achieved by dividing the support obtained by each possible tag of a variable by the maximum support for a tag of the same variable.

Advantages of the algorithm are:

- Its highly local character (each variable can compute its new consistency values given only the state at previous time step). This makes the algorithm highly parallelizable (we could have a processor to compute the new value of each variable, or even a processor to compute the consistency for each value of each variable).

- Its expressivity, since we state the problem in terms of constraints between values.

- Its flexibility, we don't have to check absolute coherence of constraints.

- Its robustness, since it can give an answer to problems without an exact solution (incompatible constraints, insufficient data...)

- Its ability to find local-optima solutions to NP problems in a non-exponential time. (Only if we have an upper bound for the number of iterations, i.e. convergence is fast or the algorithm is stopped after a fixed number of iterations. See section 4 for further details)

Drawbacks of the algorithm are:

- Its cost. Being $n$ the number of variables, $v$ the average number of possible values per variable, $c$ the average number of constraints per value, and $I$ the average number of iterations until convergence, the average cost is $n \times v \times c \times I$, that is, it depends polinomially on $n$, but for a problem with many values and constraints, or if convergence is not quickly achieved, the multiplying terms might be much bigger than $n$.

---

[2]See [Torras 89] for a clear exposition of what is relaxation labelling and what kinds of relaxation can we get by combining different support and updating functions.

- Since it acts as an approximation of gradient step algorithms, it has their typical convergence problems: Found optima may be local, and convergence is not guaranteed, since the chosen step might be too large for the function to optimize.

- In general, constraints must be written manually, since they are the modelling of the problem. This is good for easily modellable or reduced constraint-set problems, but in the case of POS tagging or WSD constraint are too many and too complicated to be written by hand.

- The difficulty to state which is the "compatibility degree" for each constraint. If we deal with combinatorial problems with an exact solution (e.g. travelling salesman), the constraints will be all fully compatible (e.g. stating that it is possible to go to any city from any other) or fully incompatible (e.g. stating that it is not possible to be twice in the same city). But if we try to model more sophisticated or less exact problems (such as POS tagging) things will not be black or white. We will have to assign a compatibility degree to each constraint.

- The difficulty to choose the support and updating functions more suitable for each particular problem.

## 3 Application to POS Tagging

In this section we expose our application of relaxation labelling to assign part of speech tags to the words in a sentence.

First we will discuss how to apply relaxation labeling techniques to POS tagging problems, and then we will propose several variations over the basic parameters: support function, updating function, compatibility values, constraint set...

Addressing tagging problems through optimization methods has been done in [Schmid 94] (POS tagging using neural networks) and in [Cowie et al. 92] (WSD using simulated annealing). Nevertheless, optimization and constraint satisfaction fields -which include relaxation labelling algorithms- are still to be paid some attention in performing NL disambiguation tasks.

The application of relaxation labelling to POS tagging has been pointed out in [Pelillo & Refice 94, Pelillo & Maffione 94], but while they use it as a toy problem to experiment their methods to improve the quality of compatibility coefficients for the constraints, we will try to apply it in a more realistic environment.

The sequence to be tagged can be of arbitrary length, even a whole corpus, but for obvious reasons, we will tag a corpus by tagging one word sequence at a time, where a word sequence is a reasonably-sized sequence of words ending in "." (reasonably-sized means small enough to fit in the memory of a computer and long enough to have several words and thus enable us to apply some constraint. Note that this definition enables a word sequence to be composed of several sentences). Previously segmented corpus or more sophisticated segmentation techniques [Briscoe 94] could have been used to extract sentences, but since constraints for POS tagging will be local enough not to connect words separated by ".", the chosen criterion seems to be enough.

The model used is the following: each word in the sentence is a variable, so we have variable 1 is the first word, variable 2 is the second one, and so on. Each variable can take several values, which are its POS tags. We need a lexicon to know them. If the lexicon doesn't contain the word, we will suppose that its possible tags are all possible POS tags for open classes.

A basic point refers to how to express the constraints: we need to state the compatibility degree of a variable value with the values of the other variables. But since length and tag position will vary from one sentence to another, we cannot state constraints in absolute terms. For instance, if we wrote $[(v_1, Determiner)(v_2, Adjective)(v_3, noun)]$ to indicate the compatibility of this three-tag sequence, then we would have to state the same compatibility for $v_2,v_3,v_4$ and for $v_4,v_5,v_6$ and all possible combinations of three successive variables.

What we need is constraints to be relative to any word, not to the sentence, for instance, we will write $[(v_i, Determiner)(v_{i+1}, Adjective)(v_{i+2}, Noun)]$ to state the compatibility of that combination in three successive words.

*The Constraint Set*

A central issue is how to obtain the constraint set. It could be built by linguistic introspection, manually stating all lexicographic and syntactical knowledge in constraint form, and then providing the result to the system, as in [Voutilainen 95]. This approach has a high labour cost, so we will try some automatic learning procedure. Usual procedures consist of extracting constraints from corpora in the form of co-occurrence statistics. Training is supervised when tagged corpora is used [Elworthy 93], while if untagged corpora is used to tune the model [Elworthy 94, Briscoe et al. 94], learning is unsupervised.

We will use supervised training, that is, computing compatibility coefficients from existing tagged corpora. In fact, we can use the same information than HMM taggers: the probability of transition from one tag to another (bigram probability) will give us an idea of how compatible they are in the positions $i$ and $i+1$, and the trigram probability will provide the same information for positions $i$, $i+1$, $i+2$. We could extend this *ad libitum*, and compute four-grams, five-grams...but computational cost would become too high, because we would have too many constraints to check for. When no evidence is available for a bigram o trigram, we will perform a rough smoothing and assign to that combination a tiny probability, just to avoid zero values which yield problems when using multiplicative functions.

We will refer to "bigrams" and "binary constraints" as the same thing, and the same applies to "trigrams" and "ternary constraints"

If we limit our constraints to binary and ternary ones, and only about adjacent words, we are wasting the power of relaxation labelling to deal with constraints between any subset of variables. So we are still interested in writing more complex constraints, such as a restriction between words $i-2$, $i$, $i+1$ and $i+4$. All relationships that are not captured by binary and ternary adjacent constraints, will be manually written. We could have used a semi-automatic learning procedure such that of [Brill 92], but we prefer to give the linguist more freedom to express any kind of constraint and not to restrict it to previously decided patterns.

So our language model will have two kinds of information: the automatically acquired (binary and ternary constraints) and the manually provided (extra and higher order constraints). On the one hand, this is an advantage, since it enables the linguist to test different models varying the constraint set, and gives a symbolic description of the model (at least of a part of it). On the other hand, it is a drawback, since we have to manually write the constraints, and what is worse, to assign a compatibility degree to each of them, without any hint about "how compatible" they are.

Accurate but costly methods to estimate compatibility values have been proposed in [Pelillo & Refice 94, Pelillo & Maffione 94]. Although our solution is not optimal like theirs, it is simpler and much cheaper computationally: We will compute the compatibility degree for the manually written constraints in the same way we did for binary and ternary ones. That is, if we estimate bigram and trigram probabilities from their occurrences in a tagged corpus, we can as well estimate constraint probabilities from the number of occurrences of the constraint pattern in the same corpus.

This is an issue that will require further attention, since as constraints can be expressed in several degrees of generality (constraints on a pair word-tag, on a tag, or on a set of tags), the occurrences (and so the estimated probabilities) in the training corpus may vary greatly for the same thing expressed with a general constraint or with a set of more specific ones.

Since we have chosen to use supervised learning, we are now depending on the availability of tagged corpora. (This could be considered another drawback). HMM models use an information we haven't used yet: the probability of a tag given a word (i.e. the prior probability of a certain tag for a word). Relaxation doesn't need it, since it is not a constraint, but it can be used to set the initial state to a not completely random one. Initially we will assign to each word its most probable tag, so we start optimization in a biassed point, hoping to be a bit nearer the maximum.

*Alternative Support Functions*

The support functions described in section 2 are traditionally used in relaxation algorithms. The most appropriate ones to our case would be (1.1), since constraint influences are added, the other functions multiply the support provided by constraints of each degree, and this means problems (multiplying zero or tiny values) when -as in our case- for a certain variable or tag no constraints are available for a given degree.

Since that functions are general, we may try to find a support function more specific for our problem. Since HMMs find the maximum sequence probability and relaxation is a maximizing algorithm, we can make relaxation maximize the sequence probability and we should get the same results. To achieve this we define a new support function, which is the sequence probability:

$$B_{ij} = \pi(v_1, t^1) \times P(v_i, t_j^i) \times (\prod_{k=1, k \neq i}^{N-1} P(v_k, t^k) \times T(t^k, t^{k+1})) \times P(v_N, t^N) \quad (3.1)$$

where
$t^k$ is the tag for variable $v_k$ with highest weight value at the current time step.
$\pi(v_1, t^1)$ is the probability for the sequence to start in tag $t^1$.
$P(v, t)$ is the lexical probability for the word represented by $v$ to have tag $t$.
$T(t_1, t_2)$ if the probability of tag $t_2$ given that the previous one is $t_1$, (in a HMM, it is the probability for the transition from $t_1$ to $t_2$).

But this function wouldn't enable us to use constraints other than bigrams, so we redefine it. So, we define the trigram contribution:

$$T_{ij} = \sum_{r \in R_{ij}^3} C_r \times p_{k_1}^{r_1} \times p_{k_2}^{r_2} \times p_{k_3}^{r_3} \quad (3.2)$$

where
$R_{ij}^3$ is the set of all ternary constraints on tag $j$ for word $i$.
and the hand-written constraints contribution:

$$C_{ij} = \sum_{r \in R_{ij}^H} C_r \times p_{k_1}^{r_1} \times \ldots \times p_{k_d}^{r_d} \quad (3.3)$$

where
$R_{ij}^H$ is the set of all hand-written constraints on tag $j$ for word $i$.

Then the new support function is:

$$S_{ij} = B_{ij} \times (1 + T_{ij}) \times (1 + C_{ij}) \tag{3.4}$$

We chose to combine the support provided by bigrams ($B_{ij}$) with the support provided by trigrams ($T_{ij}$) and hand-written constraints ($C_{ij}$) in a multiplicative form because since $B_{ij}$ is computed as the probability of the whole sequence, it is many magnitude orders smaller than $T_{ij}$ and $C_{ij}$, which are computed locally; thus, adding them would have the effect of losing the information provided by $B_{ij}$, since it would be too small to affect the other figures.

But just multiplying them yields another problem: we do not have trigram or hand written constraints for each word or tag. Then a tag with no such an information will have $T_{ij} = C_{ij} = 0$ (or, if we perform some kind of smoothing, a tiny value), and multiplying this value by $B_{ij}$ would make the support value drop. That is, a tag with trigram or hand-written constraints information would have less support than another one with only bigram information, even when the trigram information was *positive*. Since we want trigrams and other constraints to *increase* the support when positive and to *decrease* it when negative, we add one to the value before multiplying it, so when no trigrams are used, support remains unchanged, but if extra information is available, it increases/decreases the support.

*Compatibility Values*

Another point to be considered is that relaxation labelling needs constraints with a compatibility degree while we are using estimated bigram and trigram probabilities. Maybe these values are not the best ones we could be using. We try to compute a set of values that represent compatibility better than probabilities do.

Using probabilities we can compute compatibilities as:

- Mutual Information [Church & Hanks 90, Cover & Thomas 91, Ribas 94]

$$\log \frac{P(A \cap B)}{P(A) \times P(B)} \tag{4.1}$$

- Association Ratio [Resnik 93, Ribas 94]

$$P(A \cap B) \times \log \frac{P(A \cap B)}{P(A) \times P(B)} \tag{4.2}$$

- Relative Entropy [Cover & Thomas 91, Ribas 94]

$$\sum_{X \in \{A, \neg A\}, Y \in \{B, \neg B\}} P(X \cap Y) \times \log \frac{P(X \cap Y)}{P(X) \times P(Y)} \tag{4.3}$$

When considering these measures of compatibility we find that they are neither nonnegative nor limited to $[-1, 1]$ as relaxation needs. This is not a problem for (2.1) since, as described above, support values are normalized in $[-1, 1]$. But we cannot apply (2.2) to any of them since all may be negative. So we choose some normalizing functions to adequate the compatibility values to the needs of the updating function.

Although the most intuitive and direct scaling would be the linear function, we will test as well some sigmoid-shaped functions widely used in neural networks and in signal theory to scale free-ranging values in a finite interval [Kosko 90]. We will check as well the performance of (2.1) when using "normalized" compatibilities instead of free-ranging ones.

- linear 1 (to scale values in $[0, 1]$ when using (2.2))

$$\frac{1}{2}(1 + \frac{x}{\beta}) \tag{5.1}$$

- linear 2 (to scale values in $[-1, 1]$ when using (2.1))

$$\frac{x}{\beta} \tag{5.2}$$

- logistic (to scale values in $[0, 1]$ when using (2.2))

$$\frac{1}{1+e^{-2\beta x}} \tag{5.3}$$

- arc tangent (to scale values in $[-1, 1]$ when using (2.1))

$$\frac{2}{\pi}\arctan(\beta x) \tag{5.4}$$

- tanh (to scale values in $[-1, 1]$ when using (2.1))

$$\tanh(\beta x) \tag{5.5}$$

For the linear functions the $\beta$ parameter represents the range $[0, +\beta]$ or $[-\beta, +\beta]$ in which the values to be confined are distributed. Values are chosen empirically according to the kind of compatibility values used.

For the other functions, the $\beta$ parameter is the steepness of the function, although only $\beta = 1$ has been tested.

Functions (5.5) and (5.3) have the same shape, but while the former has image in $[-1, 1]$, the later has image in $[0, 1]$, so the same results could be obtained using only one of them and scaling it appropriately.

All this possibilities joined with all the possibilities of the relaxation algorithm, give a large amount of combinations and each one of them is a possible tagging algorithm.

# 4 Experiments

To this extent, we have presented the relaxation labelling algorithm family, and stated some considerations to apply them to POS tagging.

In this section we will describe the experiments performed on applying this technique to our particular problem.

We have several parameters that configure each particular relaxation algorithm.

- *Support function* We can choose among (1.1), (1.2), (1.3) and (3.4).

- *Updating function* We can choose among (2.1) and (2.2).

- *Compatibility values* We have proposed estimated probabilities, or one of (4.1), (4.2) and (4.3).

- *Confining function* It is only used when compatibility values are not probabilities, we can use (5.1), (5.3), (5.2), (5.4), (5.5), or none.

- *Constraints degree* We have binary, ternary, and hand-written constraints, we will experiment with any combination of them, as well as with a particular combination consisting of a back-off technique between trigrams and bigrams information. It will use only trigram information when available, and back-off to bigram information when not.

Our experiments will consist of tagging a corpus with all logical combinations of parameters. That is, nonsense combinations will not be tested (for instance (2.2) requires values in $[0, 1]$ so it only can be used with (5.3), (5.1) or with probabilities). The symmetric cases, i.e. (2.1) with (5.3) or probabilities, are not banned, because although values are required to be in $[-1, 1]$, and are in fact in $[0, 1]$, this doesn't mean any problems for the algorithm, it is just as we were using only constraints indicating compatibility and none indicating incompatibility.

Results at each iteration will be analyzed in order to find out if there are any significant behaviours.

Each tagging algorithm is named in the form S$w$A$x$V$y$F$z$, where:

- $w$ indicates the support function (**s** for (1.1), **p** for (1.2), **m** for (1.3), **q** for (3.4)).

- $x$ indicates the updating function (**c** for (2.1), **p** for (2.2)).

- $y$ indicates which kind of values are used (**p** for probabilities, **i** for (4.1), **k** for (4.2), **h** for (4.3)).

- $z$ indicates the confining function used (**l** for (5.2) and (5.1), **s** for (5.3), **t** for (5.4), **h** for (5.5), **n** for none, no F$z$ appears when using probabilities, since no scaling is needed).

**B**,**T**,**C**,**K** are added to the name to indicate which constraint information is being used (**B**igrams, **T**rigrams, hand-written **C**onstraints, or Bac**K**-off), Several of them are allowed (e.g. **B** and **C** at the same time) but **K** is only allowed with **C**.

So for instance **SpAcViFhBT** is the algorithm obtained with support function (1.2), updating function (2.2), using values (4.1) confined with (5.5) and using bigram and trigram information.

Four experiments have been performed: the first over a "toy" corpus in Spanish, two over large English corpora, and the last one over a big Spanish corpus.

In order to have a comparison reference we will evaluate the performance of two taggers: A blind most-likely-tag tagger and a HMM tagger [Elworthy 93] performing Viterbi algorithm . The train and test corpora will be the same for all taggers.

All figures are given over ambiguous words, (computing it over all the words yields higher figures, but not better algorithms).

The support functions described above are applied considering that the possible degrees of constraints are 2,3, and more than 3. That is, binary and ternary constraints are computed as stated by (1.1), (1.2) and (1.3), but hand written constraints are treated as a unique set of constraints, without distinctions of degree or involved variables. This is done to ease and speed the computations.

## 4.1 First experiment

The first experiment is performed on a "toy" Spanish corpus.

**Corpus** : "El sur" a novel by Adelaida Garcia Morales, some 17Kw, tagged with a 70-tag[3] tag set [Moreno-Torres 94].
**Training set** : 15 Kw.
**Test set** : 2 Kw randomly extracted.

Results:
        Most-likely : 69.62% correctly tagged ambiguous words.
        HMM       : 94.62% correctly tagged ambiguous words.

        Relaxation : 95.38% correctly tagged ambiguous words.

---

[3]see Appendix A.1

This result for relaxation is the best result obtained when using only the same information than a HMM tagger (i.e. lexical probabilities and bigrams). It is obtained with $SsApViFsB$, that is, with support function (1.1), updating function (2.2), and converting probabilities to mutual information ratios (4.1), confined to $[0, 1]$ with (5.3).

But relaxation can deal with more constraints, so we added a set of some 50 hand-written constraints, obtained adapting the context constraints used in [Moreno-Torres 94][4]. Those constraints were derived analyzing the most frequent errors committed by the tagger exposed in that work.

The constraints do not intend to be a general language model, they cover only some common error cases, and are not exhaustive. This will cause that tagging using only hand-written constraints (no bigrams information) produces rather bad results, since not enough information is available for the relaxation algorithm. So, experiments with only hand-written constraints are not performed.

The compatibility degree for these constraints is computed from their occurrences in the corpus.

Best result is 95.77% obtained in the same above case plus constraints: $SsApViFsBC$, and with $SsApViFlBC$ ($SsApViFlB$ gave 94.62%, the same than HMM).

We get that relaxation labelling may outperform HMM in the same conditions (i.e. using the same information), when using the adequate functions. But what is more important, results can be improved feeding the algorithm with additional constraints

Next we give the results obtained using ternary constraints, in different algorithm combinations.

Best results using ternary constraints:
    using only trigrams: 90.00%
    using bigrams and trigrams: 94.23%
    using trigrams and hand-written: 92.31%
    using bigrams, trigrams and hand-written: 94.62%

That is, when combining trigrams and bigrams, results obtained are worse than when using bigrams alone, and trigrams alone are far from achieving interesting results. This is probably due to the fact that the training corpus is small (15Kw) and there is little evidence to collect a good trigram model (in fact, we collected a model which misleads the algorithm instead of helping it). We will contrast this issue later in the experiments over bigger corpora.

---

[4] see Appendix B.1

*Searching a more specific support function.*

We have been using support functions that are traditionally used in relaxation, but we might try to find a more appropriate one to our case, that is, we can try to specialize relaxation labelling to POS tagging.

The goodness of the results achieved by relaxation depends on the support function which it is maximizing. If the support function were the probability of the tag sequence, it would be computing the same than Viterbi algorithm [Viterbi 67, Baum 72] (see also [Cutting et al. 92] for a clear exposition) and thus, we should get the same results. If we add to this hand-written constraints, could we improve the results as we did before ?

Results obtained with this specific support function (3.4) are the following:

Best result using only bigrams: 94.23% with $SqApVkFnB$ updating function (2.2) and using association ratio values.
The same result is obtained with all combinations *B, BT, BC, BTC*

We are maximizing the sequence probability, just like Viterbi does, but we get slightly worse results. There are two main reasons for that. The first one is that relaxation does not maximize the support function (the sequence probability in this case) but the *weighted* support for each variable, so we are not doing exactly the same than a HMM tagger. Second reason is that relaxation is not an algorithm that finds global optima and can be trapped in local maximums, while Viterbi computes the globally maximum probability sequence.

Results for trigrams alone are not given, since the support function (3.4) definition is based on transition probabilities, so it makes no sense trying to compute that support value without using bigrams.

Using trigrams or hand-written constraints doesn't make any difference.

*Stopping before convergence.*

All above results are obtained stopping the relaxation algorithm when it reaches convergence (no significant changes are produced from one iteration to the next), but relaxation algorithms not necessarily give their best results at convergence[5],[Lloyd 83, Richards et al. 81] or not always need to achieve convergence to know what the result will be [Zucker et al. 81]. So they are often stopped after a few iterations. Actually, what we are doing is changing our convergence criterion to one more sophisticated than "stop when there are no more changes".

---

[5]This is due to two main reasons: (1)The optimum of the support function doesn't correspond *exactly* to the best solution for the problem, that is, the chosen function is only an approximation of the desired one. And (2) performing too much iterations can produce a more probable solution, which will not necessarily be the correct one.

In our case, experiments show that in certain cases results after the first few iterations are better than at convergence, while in other cases best policy is waiting for convergence. We distinguished four behaviour patterns: (1) best result at the very first iterations, (2) best results after some ten iterations, (3) best result after some twenty iterations , (4) best result at convergence.

Some of the results obtained are summarized in tables 1 and 2. In table 1 best overall results are shown, and table 2 shows samples of all behaviour patterns. Each column contains the best result obtained in the iteration range specified.

| Algorithm | it.1-3 | it.9-11 | it.18-20 | conv. | |
|---|---|---|---|---|---|
| SsApViFsBTC | 94.62% | 94.62% | 94.62% | 94.62% | (best at 1-3) |
| SsApViFsBC | 94.23% | 95.77% | 95.77% | 95.77% | (best at 9-11, 18-20 and conv.) |

Table 1: Best results

| Algorithm | it.1-3 | it.9-11 | it.18-20 | conv. | |
|---|---|---|---|---|---|
| SpApViFsBC | 93.46% | 92.69% | 92.69% | 92.69% | behaviour (1) |
| SsAcVkFhBTC | 92.69% | 93.85% | 93.46% | 92.69% | behaviour (2) |
| SsApViFlBTC | 92.31% | 93.46% | 94.23% | 93.85% | behaviour (3) |
| SsApViFlBC | 90.00% | 95.38% | 95.38% | 95.77% | behaviour (4) |

Table 2: Behaviour patterns samples

Another point for further studies must be which stopping criterions are adequate for different relaxation algorithms. Some heuristic criterions have been proposed in [Haralick 83], but they have not been tested yet in our case.

*Combining information in a Back-off hierarchy.*

We defined at the beginning of section 4 a back-off mechanism to combine bigram and trigram information: Use trigrams if available and bigrams when not. Results obtained with this method are comparable -or worse- to the ones obtained with trigrams alone. The cause of these bad results is the same than above: The trigram model is based on too small evidence. We'll get back to this issue when dealing with bigger corpora.

## 4.2 Second experiment

The second experiment is performed over a small English corpus

**Corpus** : Susanne corpus, about 147 Kw, tagged with a 150-tag tag set[6].
**Training set** : 141 Kw.
**Test set** : 6 Kw randomly extracted.

Results:
      Most-likely : 86.01% correctly tagged ambiguous words.
      HMM      : 93.20% correctly tagged ambiguous words.

      Relaxation : 92.08% correctly tagged ambiguous words.

This result is obtained with support function (1.1), updating function (2.2), and values (4.1) linearly normalized.

In this case, the results obtained by relaxation are worse that those of a HMM tagger, when using the same information, and waiting for convergence.

But in this corpus relaxation has a stronger tendency to give its best results before convergence that it had in the small Spanish corpus. So, the best result obtained by relaxation using only bigrams is 93.12%, obtained by $SpApViFlB$ at iteration 18.

This is still slightly worse than the HMM tagger. Let's see what happens when using extra information.

Best results using ternary constraints:
      using only trigrams: 93.16%
      using bigrams and trigrams: 93.58%
      using trigrams and hand-written: 93.62%
      using bigrams, trigrams and hand-written: 93.70%

The use of trigrams improves the results, and raises up to 93.58% in iteration 6 of $SsAcViFsBT$ or 93.54% at iteration 2 of $SsApViFsBT$

Using trigrams alone is not far below using bigrams alone, giving 93.16% at iteration 7 of $SsAcViFsT$.

It is interesting to note that in the small Spanish corpus, the use of trigrams didn't represent an improvement (in fact it was misleading) but here trigrams do help the algorithm, and obtained results are better when adding trigrams information. This is due to the fact that the Spanish corpus was to small

---

[6]see Appendix A.2

to collect a reasonable trigram model, and this corpus seems to have a more adequate size.

If we analyze the errors committed by the HMM tagger, on the test corpus, we will see that an important part of them are caused by a tagging convention adopted in susanne corpus: locutions such as "as well as" or "on the part of" are tagged as a whole: "as_CC well_CC as_CC" and "on_II the_II part_II of_II". This introduces noise, since words like "the" that are usually unambiguous can have here several tags. This may also contribute to lower the results of HMM taggers.

Although the errors committed by the relaxation algorithm are not necessarily the same than those committed by the HMM tagger, we may assume that these errors are the main weakness of our model, and since we have a flexible tool such as relaxation, we can try to write the necessary constraints to solve these ambiguities.

Using a set of 66 constraints[7] -64 of them corresponding to locutions-, the performance is increased up to 93.70% in iteration 5 of $SsAcViFsBTC$ or in iteration 19 of $SsApViFlBC$. Other good results are 93.62% in iteration 4 of $SsAcVkFlTC$ and in iteration 2 of $SsApViFsBTC$.

*Searching a more specific support function.*

The use of support function (3.4) gives -as in the previous case- results which are slightly below those given by the HMM tagger.

Best result is 92.31% obtained by $SqApVpBC$.
$SqApVpBC$ obtains 92.19%.

This and other results seem to indicate that in this case, the use of constraints and/or trigrams improves slightly performance, while in the previous experiment it didn't.

*Stopping before convergence.*

The same behaviours than before can be found in this case, but as mentioned above, there is a stronger tendency to reach the optimum before convergence, maybe due to the fact that the model contains the locutions tagged as whole, and trying to get more probable taggings leads us to have worse solutions.

Results obtained are summarized in tables 3 and 4. In table 3 best overall results are shown, and table 4 shows samples of all behaviour patterns.

---

[7]see Appendix B.2

| Algorithm | it.1-3 | it.9-11 | it.18-20 | conv. | |
|---|---|---|---|---|---|
| SsApViFsBTC | 93.62% | 92.15% | 90.49% | 87.40% | (best at 1-3) |
| SsAcViFsBTC | 92.11% | 93.62% | 92.42% | 88.94% | (best at 9-11) |
| SsApViFlBC | 90.68% | 92.66% | 93.70% | 93.12% | (best at 18-20 and conv.) |

Table 3: Best results

| Algorithm | it.1-3 | it.9-11 | it.18-20 | conv. | |
|---|---|---|---|---|---|
| SsApViFsBTC | 93.62% | 92.15% | 90.49% | 87.40% | behaviour (1) |
| SsAcViFsBTC | 92.11% | 93.62% | 92.42% | 88.94% | behaviour (2) |
| SsApViFlBC | 90.68% | 92.66% | 93.70% | 93.12% | behaviour (3) |
| SpApVkFlBC | 87.48% | 89.80% | 90.57% | 92.19% | behaviour (4) |

Table 4: Behaviour patterns samples

*Combining information in a Back-off hierarchy.*

In this experiment results obtained with a Back-off technique between bigrams and trigrams are similar to those obtained with trigrams alone: we get 92.42% at iteration 4 of *SsAcVkFnK* and 93.08% at the same iteration of *SsAcVkFnKC*.

## 4.3 Third experiment

The third experiment is performed over a big English corpus

**Corpus** : Preliminary version of Wall Street Journal, about 1061 Kw, tagged with a 45-tag tag set[8].
**Training set** : 1055 Kw.
**Test set** : 6 Kw randomly extracted.

Results:
       Most-likely : 88.52% correctly tagged ambiguous words.
       HMM      : 93.63% correctly tagged ambiguous words.

       Relaxation : 92.72% correctly tagged ambiguous words.

This result is obtained, as well as in Susanne corpus- with support function (1.1), updating function (2.2), and values (4.1) linearly normalized.

These are the results obtained by relaxation are worse that those of a HMM tagger, when using the same information, and waiting for convergence.

---
[8]see Appendix A.3

This experiment is similar to the previous one, in the way that relaxation has a stronger tendency to give its best results before convergence that it had in the Spanish corpus. So, the best result obtained by relaxation using only bigrams is 93.85%, obtained by $SpApViFlB$ at iteration 18.

This is already better than the HMM tagger. Let's see what happens when using extra information.

Best results using ternary constraints:
    using only trigrams: 94.07%
    using bigrams and trigrams: 94.04%
    using trigrams and hand-written: 94.10%
    using bigrams, trigrams and hand-written: 94.01%

In this case we can appreciate that the use of trigrams yields an improvement much clearer than in previous experiments. This is due to the fact that the training corpus is really big here, so the trigram model obtained is much better than in previous cases.

As in the previous experiment, we analyzed the errors most frequently committed by the HMM tagger, and considered these to be the weak point of our model. We wrote constraints to solve some of these errors.

Using a set of 31 constraints[9] the performance is increased up to 94.10% in iteration 9 of $SmApViFlTC$ or to 94.07% in iteration 4 of $SmAcViFsTC$.

*Searching a more specific support function.*

The use of support function (3.4) gives -as in the previous cases- results which are slightly below those given by the HMM tagger.

Best result is 93.60% obtained by $SqApViFsB$.

The use of trigrams or hand written constraints doesn't improve performance in a significative way.

*Stopping before convergence.*

The same behaviours than before can be found in this case, but like in Susanne corpus, there is a tendency to reach the optimum before convergence. Whether this is due to the fact that English experiments are using a model obtained from bigger corpora than the Spanish experiment is a matter that must be further studied, that is, whether the goodness or generality of the model influences or not the number of iterations needed to achieve the optimum.

Results obtained are summarized in tables 5 and 6. In table 5 best overall results are shown, and table 6 shows samples of all behaviour patterns.

---

[9] see Appendix B.3

| Algorithm   | it.1-3  | it.9-11 | it.18-20 | conv.  |                  |
|-------------|---------|---------|----------|--------|------------------|
| SmAcViFsK   | 94.04%  | 92.51%  | 89.37%   | 86.64% | (best at 1-3)    |
| SmApViFlTC  | 93.35%  | 94.10%  | 93.41%   | 89.09% | (best at 9-11)   |
| SsApViFlBTC | 91.31%  | 93.51%  | 93.89%   | 92.16% | (best at 18-20)  |
| SsApVkFlBTC | 90.03%  | 91.13%  | 91.72%   | 92.35% | (best at conv.)  |

Table 5: Best results

| Algorithm   | it.1-3  | it.9-11 | it.18-20 | conv.  |               |
|-------------|---------|---------|----------|--------|---------------|
| SmAcViFsK   | 94.04%  | 92.51%  | 89.37%   | 86.64% | behaviour (1) |
| SmApViFlTC  | 93.35%  | 94.10%  | 93.41%   | 89.09% | behaviour (2) |
| SsApViFlBTC | 91.31%  | 93.51%  | 93.89%   | 92.16% | behaviour (3) |
| SsApVkFlBTC | 90.03%  | 91.13%  | 91.72%   | 92.35% | behaviour (4) |

Table 6: Behaviour patterns samples

*Combining information in a Back-off hierarchy.*

Although it does not appear in the previous tables because they show results at fixed iterations, the best overall result is 94.26% obtained in iteration 4 of *SmAcViFsK* and *SmAcViFsKC*. That is, the back-off technique proposed earlier seems to work when using a trigram model build from enough evidence.

## 4.4 Fourth experiment

The last experiment is performed over a Spanish corpora tagged using the morphological analyzer described in [Acebo et al. 94] and the tagger descibed in [Moreno-Torres 94]. See [Cervell et al. 95] for further details.

It is a 570 Kw corpus, of which 23 Kw are hand-tagged. This part are Spanish press articles and are what we used in the experiments.

Results are preliminar since the corpus is just built and many items are to be verified.

**Corpus** : 23 Kw, tagged with a 70-tag tag set[10].
**Training set** : 17 Kw.
**Test set** : 6 Kw randomly extracted.

Results:
        Most-likely : 90.09% correctly tagged ambiguous words.
        HMM      : 90.99% correctly tagged ambiguous words.

        Relaxation : 90.50% correctly tagged ambiguous words.

---
[10]see Appendix A.4

This is the result obtained at convergence by $SsAcVhFsB$.

It is surprising that the most-likely algorithm yields a 90.09% on ambiguous words, and that more informed algorithms don't improve it much. Further analysis must be done on the tagging of the corpus and on the tagset to find out the causes.

Nevertheless, relaxation can once more improve performance over the HMM tagger, when using extra information:

Adding trigrams doesn't improve the results, since as in first experiment we are using a corpus too small to collect a good trigram model.

If we adapt some of the constraint used in the first Spanish corpus[11], we reach at convergence 91.08% with $SsAcVhFsBC$.

As in previous cases, best overall results are not obtained at convergence: we get 92.26% at iteration 4 of $SsAcViFsB$, improved to 92.51% when using hand-written constraints with the same algorithm ($SsAcViFsBC$).

## 5 Application to word sense disambiguation

Word sense disambiguation (or word sense selection) consists of, given a sentence, assigning to each word a sense label, which can be a pointer to a sense entry for that word in a MRD or in a word taxonomy, or a more general label, such a "topic" for that word in that context. Methods used to do this vary greatly. The knowledge used may be obtained from lexicographer introspection [Hirst 87], from existing Machine Readable Dictionaries (MRDs) [Rigau 94, Harley 94], thesaurus or taxonomies (such as WordNet [Miller et al. 91]), or from statistical processing of tagged/untagged corpora. This later approach includes lexical statistics (computing mutual information, relative entropy, or merely frequencies of words and senses. [Miller et al. 93]), and lexical distributions (computing and comparing distribution of senses respect to a context, which may be global [Yarowsky 92] or local [Yarowsky 93, Bruce & Wiebe 94]). See [Resnik 93, Ribas 95] for clear overviews.

Results vary from 60% (lower bound established by [Miller et al. 94]) to 90%-95% (like in [Yarowsky 93, Bruce & Wiebe 94]) depending on the kind of knowledge used (statistical, taxonomy...), the granularity of labels assigned (ranging from coarse-grained like topic identifiers or word classes in a taxonomy to more fine-grained such as dictionary/thesaurus entries or taxonomy nodes) and the performed experiments (which are usually done over a reduced set of words). Although this variability makes very difficult to establish a comparison between different methods, some steps have been done in this direction in [Gale et al. 92].

---

[11] see B.4

The relaxation labelling algorithm can be used to select noun senses in corpora.

Preliminary experiments have been performed over a corpus of triples {*verb, syntactic position, noun*} extracted from a parenthesized corpus. A sentence is disambiguated as a whole, being a sentence the set of all triples related to the same verb.

This approach is chosen since having the corpus analyzed must help the disambiguation. Since this is a prospective experiment, we tried it the easy way.

Nevertheless, starting from a set of triples extracted from a parenthesized corpus is seldom possible. So our efforts will address in the future WSD (or WSD combined with POS tagging) on plain text corpora.

Our variables will be the words, and their possible values will be an identifier of a sense in a concept taxonomy. We chose for this WordNet [Miller et al. 91].

We will need constraints such as $[verb, position, sense]$ (a syntactic position of a certain verb tends to prefer/reject a kind of sense) those are the "binary constraints". (We could also express them depending on verb senses, but the corpus used to train it didn't include that information).

And even we could imagine more complex constraints such as $[verb, (position_1, sense_1)(position_2, sense_2)]$ (for instance, to express the higher compatibility of a certain kind of subject with a special kind of object).

It is harder to obtain the constraints automatically in this case, since we would require to compute the probability of each possible sense given each possible verb !!

What we need are more general constraints, i.e. class-based constraints, stating that a certain verb -or kind of verb- tends to have subjects of a certain type, without having to enumerate all the concepts of that type, so we would say $[eat, subj, ANIMATE]$ to assess the compatibility between the verb "eat" and a type of subject, and we wouldn't need to express it for each animate thing.

To obtain these constraints automatically means entering in selectional restrictions extraction [Resnik 93, Ribas 95], and this is not the aim of this paper. In our work we used a bigram table to sense level as binary constraints and general hand-written constraints to check special cases. Obviously the bigram table was computed only for verbs and senses appearing in the corpus.

The use of general constraints requires a slight modification of the relaxation algorithm, since now constraints must be applied not only to the sense that they express (ANIMATE in the previous example) but also to all its hyponyms in the taxonomy (anything that is animated).

Results obtained are not significant due to the size of the corpus and the particularity of tested cases, but they are encouraging, since suggest that the use of adequate constraints and relaxation algorithms can perform a good disambiguation.

# 6 Conclusions

We have applied relaxation labelling algorithm to the task of POS tagging. Results obtained show that the algorithm not only can equal markovian taggers, but also outperform them when given enough constraints or a good enough model.

We can state that in all experiments, the refinement of the model with hand written constraints led to an improvement in performance. We obtained a neat improvement in performance adding few constraints which were not linguistically motivated. Probably adding more and more "linguistic" constraints would yield more significant improvements.

It seems clear than relaxation can deal with complex models of POS sequences, and that these models can be improved or tunned adding or changing their constraints.

Several parametrizations for relaxation have been tested, and although it is not clear whether best results are obtained with support function (1.1) or with (1.3), it seems that (1.2) gives poorer results.

The results obtained also point out that the best policy is using mutual information as compatibility values.

It is not clear which are the most appropriate updating and normalizing functions.

Further experiments and theoretical work must be done to assess which are the most appropriate parameters and functions for NLP.

A general conclusion of this work is that optimization algorithms may be adequate to deal with NLP tasks. All what we need is to find the function which must be optimized, and then select the most appropiate method to do it.

*Searching a more specific support function.*

Results obtained using support function (3.4) are in all cases below those of the HMM tagger. This suggests that this function has still to be improved to become an adequate support function for POS-tagging.

The fact that using extra information (trigrams and hand-written constraints) doesn't improve performance significatively i two of the three experiments suggests that this function doesn't combine adequately bigram, trigram and constraints information, and that it relies too much on bigram information.

Further work is required on this issue.

*Stopping before convergence.*

Most relaxation algorithms used seem to give their best results much before reaching convergence. Many of them have troubles in reaching convergence and must be stopped to avoid too long computations.

This suggests that waiting for convergence (no more significant changes) is not a good policy, and that alternative stopping criterions must be studied.

*Combining information in a Back-off hierarchy.*

The results obtained by the back-off technique are only better than the others in the third experiment. The same could be said about the use of trigrams.

This is probably due to the fact that the third experiment is the only one with a really big training corpus, which produces a good trigram model, while the other experiments do not have enough evidence to collect a good model.

# 7 Future work

Much work is still to be done, ranging from the appropriateness of relaxations algorithms (better support and confining functions, convergence criterions, etc.) to the possible combination of POS and WSD problems, through the quality of constraints used (automatic acquisition, choosing a compatibility degree, etc.),

Some important issues are:

- Experiment stochastic relaxation (Simulated annealing).

- Model the WSD problem and the combined problem POS-tagging plus WSD, using crossed information to disambiguate both kinds of tags. This should be done on non-analyzed corpora.

- Experiment starting points other that most-likely assignment (e.g. random starting point, start from the combination given by a HMM tagger, etc.)

- Consider automatically extracted constraints.

- Investigate alternative ways to compute compatibility degrees for hand-written constraints.

- Study back-off techniques that take into account all classes and degrees of constraints and allow assigning them different influence values.

- Experiment with different convergence and stopping criterions.

- Study other support functions.

- Study the confining functions and values used as compatibilities and the possible alternatives.

- Compare other optimization or constraint satisfaction techniques (gradient step, neural networks, genetic algorithms, ...) applied to NLP tasks.

# 8 Acknowledgements

I thank Horacio Rodríguez for his valuable comments on this paper. I also thank Kiku Ribas, German Rigau and Pedro Meseguer for their interesting suggestions.

# A  Tag sets

## A.1  Spanish novel corpus

| | | |
|---|---|---|
| A   | Pa  | Tc  |
| Cc  | Pc  | Td  |
| Cq  | Pd  | Te  |
| Cs  | Pi  | Ti  |
| Da  | Pl  | To  |
| Dc  | Pm  | Tr  |
| Dg  | Po  | Ts  |
| Do  | Pq  | Tx  |
| Ds  | Pr  | U   |
| Dv  | Ps  | V0  |
| Dx  | Pv  | V1  |
| E0  | Px  | V2  |
| E1  | Ra  | V3  |
| E2  | Rb  | W   |
| E3  | Rd  | X   |
| H0  | Re  | Z   |
| H1  | Rl  | Z   |
| H2  | Ro  | Z!  |
| H3  | Rp  | Z"  |
| J   | Rs  | Z,  |
| Nc  | Rv  | Z-  |
| No  | S   | Z.  |
| O   | Ta  | Z?  |

## A.2  Susanne corpus

| | | | |
|---|---|---|---|
| !    | ICS  | NP2   | VBDR |
| $    | IF   | NPD1  | VBDZ |
| &FO  | II   | NPD2  | VBG  |
| &FW  | IO   | NPM1  | VBM  |
| (    | IW   | PN    | VBN  |
| )    | JA   | PN1   | VBR  |
| ,    | JB   | PNQO  | VBZ  |
| .    | JBR  | PNQS  | VD0  |
| ...  | JBT  | PNQVS | VDD  |
| :    | JJ   | PP$   | VDG  |
| ;    | JJR  | PPH1  | VDN  |
| ?    | JJT  | PPHO1 | VDZ  |
| APP$ | LE   | PPHO2 | VH0  |
| AT   | MC   | PPHS1 | VHD  |
| AT1 | MC1 | PPHS2 | VHG |
| BTO | MC2 | PPIO1 | VHN |
| CC | MD | PPIO2 | VHZ |
| CCB | MF | PPIS1 | VM |
| CS | ND1 | PPIS2 | VMK |
| CSA | NN | PPX1 | VV0 |
| CSN | NN1 | PPX2 | VVD |
| CST | NN2 | PPY | VVG |
| CSW | NNJ | RA | VVGK |
| DA | NNJ1 | REX | VVN |
| DA1 | NNJ2 | RG | VVNK |
| DA2 | NNL | RGA | VVZ |
| DA2R | NNL1 | RGQ | XX |
| DAR | NNL2 | RGQV | ZZ1 |
| DAT | NNO | RL | g |
| DB | NNS | RP | |
| DB2 | NNS1 | RPK | |
| DD | NNS2 | RR | |
| DD1 | NNSA1 | RRQ | |
| DD2 | NNSB2 | RRQV | |
| DDQ | NNT1 | RRR | |
| DDQ$ | NNT2 | RRT | |
| DDQV | NNU | RT | |
| EX | NNU1 | TO | |
| FA | NNU2 | UH | |
| FB | NP1 | VB0 | |

## A.3 WSJ corpus

| | | |
|---|---|---|
| # | JJR | RP |
| $ | JJS | SYM |
| '' | LS | TO |
| ( | MD | UH |
| ) | NN | VB |
| , | NNS | VBD |
| . | NP | VBG |
| : | NPS | VBN |
| CC | PDT | VBP |
| CD | POS | VBZ |
| DT | PP | WDT |
| EX | PP$ | WP |
| FW | RB | WP$ |
| IN | RBR | WRB |
| JJ | RBS | `` |

## A.4 Spanish press corpus

| | | |
|---|---|---|
| A | M | VOEV |
| COA | N | VOHV |
| COC | POA | VOHVA |
| COS | POD | VOP |
| D | POI | VOS |
| GOEP | POL | VOV |
| GOEV | PON | W |
| GOHP | POO | Y |
| GOHV | POP | ZO |
| GOP | POQ | ZO |
| GOS | POR | ZO! |
| GOV | POS | ZO, |
| I | ROA | ZO- |
| IOE | TOD | ZO. |
| IOEP | TOI | ZO; |
| IOEV | TOO | ZO? |
| IOH | TOP | ZOX |
| IOHP | TOQ | |
| IOHS | UOEV | |
| IOHV | UOHV | |
| IOP | UOP | |
| IOS | UOV | |
| IOV | VOE | |
| J | | |

# B   Hand written constraints

Constraints have two main parts: the heart, closed in angle brackets ¡¿, and the body (the rest of the constraint).

The heart is the word or tag (or pair word tag) which will have its support affected by the constraint when the words around it match the body of the constraint.

Each word position in the constraint body may be a word, a tag, or a set of tags closed in square brackets [].

A certain flexibility in positions is allowed by using the star character *. For instance,

- \* (or *1..1) means that any word/tag can be in that position, but that there must be one.

*0..3 means that there can be from 0 to 3 words whatever they are. Backtracking is performed to adjust the number of words that must be skipped. Shortest constraint match is taken.

*2..2 means that there must be exactly two words, whatever they are.

## B.1   Spanish novel corpus

```
\Z,\ <\A\> \Ra\;
\Z,\ <\V0\> \Ra\;
"tal" * <\Cq\>;
<\Cq\> \V2\;
<\Cq\> \H2\;
<\Cq\> \E2\;
\Ta\ *0..8 <\Cq\>;
\To\ * <\Cq\>;
"despus" \Rd\ <\Cq\>;
"antes" \Rd\ <\Cq\>;
\Cq\ *0..8 [\Cc\ \Z,\] <\Cq\>;
[\E0\ \E1\ \E2\ \E3\] \A\ <\Cq\>;
[\V0\ \V1\ \V2\ \V3\] \A\ <\Cq\>;
"ms" *1..3 <"que",\Cq\>;
\Z.\ \A\ <\Cq\>;
[\Do\ \To\ \Pm\ \Po\] *0..9 <\Cq\>;
"sino" <"que",\Cq\>;
[\Z\ \Z\ \Z!\ \Z"\ \Z,\ \Z-\ \Z.\ \Z?\] <"sino",\Cs\>;
[\Z\ \Z\ \Z!\ \Z"\ \Z,\ \Z-\ \Z.\ \Z?\] <"Sino",\Cs\>;
"no" *0..9 <"sino",\Cs\>;
"tanto" <"como", \Cs\>;
```

```
[\E0\ \V0\] *0..5 <"as",\Ds\>;
[\E0\ \V0\] *0..5 <"as",\V0\>;
<\Dv\> [\Z\ \Z\ \Z!\ \Z"\ \Z,\ \Z-\ \Z.\ \Z?\];
\V0\ <\E2\>;
<\J\> \To\;
\V0\ <\Pc\> [\Z\ \Z\ \Z!\ \Z"\ \Z,\ \Z-\ \Z.\ \Z?\];
<\Pc\> [\Ra\ \Rb\ \Rd\ \Re\ \Rl\ \Ro\ \Rp\ \Rs\ \Rv\];
<\Pm\> [\Ra\ \Rb\ \Rd\ \Re\ \Rl\ \Ro\ \Rp\ \Rs\ \Rv\];
<\Pm\> [\Z\ \Z\ \Z!\ \Z"\ \Z,\ \Z-\ \Z.\ \Z?\];
<\Pm\> \V0\;
\Pm\ \Cc\ <\Pm\>;
\Pm\ \Cc\ <\Do\>;
\S\ [\V1\ \Pd\ \A\ \Ps\ \Re\ \Rd\] <\Pq\>;
\S\ [\Ta\ \Tc\ \Td\ \Te\ \Ti\ \To\ \Tr\ \Ts\ \Tx\] <\Pq\>;
\Pq\ *0..8 "pero" <\Pq\>;
"cosa" *0..2 <"que", \Pq\>;
"Cosa" *0..2 <"que", \Pq\>;
[\Pv\ \Pc\ \Td\] *0..5 <"que", \Cq\>;
[\Pv\ \Pc\ \Td\] *0..5 <"que", \Pq\>;
\Pq\ *0..7 \Cc\ <\Pq\>;
[\To\ \Tc\] <\S\>;
"ms" <\S\> "que";
<\Ta\> \Po\;
<\Tc\> \Ts\;
[\Ra\ \Rb\ \Rd\ \Re\ \Rl\ \Ro\ \Rp\ \Rs\ \Rv\]
   \S\
   <\Tc\>
    [\Z\ \Z\ \Z!\ \Z"\ \Z,\ \Z-\ \Z.\ \Z?\];
\S\ \A\ <\Tc\> [\Z\ \Z\ \Z!\ \Z"\ \Z,\ \Z-\ \Z.\ \Z?\];
[\Z\ \Z\ \Z!\ \Z"\ \Z,\ \Z-\ \Z.\ \Z?\] <\Tc\> \Ds\;
<\Td\> [\A\ \Nc\ \Tc\];
\S\ <\Td\> \Pq\;
\S\ <\Ts\> [\Pq\ \E0\ \E1\ \E2\ \E3\];
<"s",\V0\> *0..1 [\Ti\ \Pi\];
[\H0\ \H1\ \H2\ \H3\] \E1\ <\V1\>;
```

## B.2  Susanne corpus

```
<"out",\II\> "of";
"out" <"of",\II\>;
<"of",\RR\> "course";
"of" <"course",\RR\>;
<"once",\RR\> "again";
"once" <"again",\RR\>;
```

```
<"along",\II\> "with";
"along" <"with",\II\>;
<"such",\II\> "as";
"such" <"as",\II\>;
<"next",\II\> "to";
"next" <"to",\II\>;
<"because",\II\> "of";
"because" <"of",\II\>;
<"except",\II\> "for";
"except" <"for",\II\>;
<"such",\II\> "as";
"such" <"as",\II\>;
<"away",\II\> "from";
"away" <"from",\II\>;
<"up",\II\> "to";
"up" <"to",\II\>;
<"a",\DD\> "lot";
"a" <"lot",\DD\>;
<"the",\DD\> "rest";
"the" <"rest",\DD\>;
<"even",\CS\> "though";
"even" <"though",\CS\>;
<"not",\LE\> "only";
"not" <"only",\LE\>;
<"at",\RR\> "last";
"at" <"last",\RR\>;
<"in",\II\> "front" "of";
"in" <"front",\II\> "of";
"in" "front" <"of",\II\>;
<"with",\II\> "respect" "to";
"with" <"respect",\II\> "to";
"with" "respect" <"to",\II\>;
<"with",\II\> "regard" "to";
"with" <"regard",\II\> "to";
"with" "regard" <"to",\II\>;
<"in",\II\> "terms" "of";
"in" <"terms",\II\> "of";
"in" "terms" <"of",\II\>;
<"by",\II\> "means" "of";
"by" <"means",\II\> "of";
"by" "means" <"of",\II\>;
<"by",\RR\> "the" "way";
"by" <"the",\RR\> "way";
"by" "the" <"way",\RR\>;
```

```
<"as",\CC\> "well" "as";
"as" <"well",\CC\> "as";
"as" "well" <"as",\CC\>;
<"as",\CSA\> \JJ\ "as";
"as" <\JJ\> "as";
"as" \JJ\ <"as",\CSA\>;
<"on",\II\> "the" "part" "of";
"on" <"the",\II\> "part" "of";
"on" "the" <"part",\II\> "of";
"on" "the" "part" <"of", \II\>;
<"one",\DD1\> "and" "the" "same";
"one" <"and",\DD1\> "the" "same";
"one" "and" <"the",\DD1\> "same";
"one" "and" "the" <"same", \DD1\>;
[\VBZ\ \VB0\ \VHZ\ \VHD\ \VH0\] *0..5 <\VVN\>;
[\VBZ\ \VB0\ \VHZ\ \VHD\ \VH0\] *0..5 <\VVD\>;
```

## B.3   WSJ corpus

```
<"as",\RB\> * "as" ;
"too" <\RB\> ;
"too" <\JJ\> ;
"too" <\VBG\> ;
"too" <\VBN\> ;
"have" *0..4 <\VBN\> ;
"have" *0..4 <\VBD\> ;
"has" *0..4 <\VBN\> ;
"has" *0..4 <\VBD\> ;
"had" *0..4 <\VBN\> ;
"had" *0..4 <\VBD\> ;
"be" *0..4 <\VBN\> ;
"be" *0..4 <\VBD\> ;
"been" *0..4 <\VBN\> ;
"been" *0..4 <\VBD\> ;
"is" *0..4 <\VBN\> ;
"is" *0..4 <\VBD\> ;
"''s" *0..4 <\VBN\> ;
"''s" *0..4 <\VBD\> ;
"are" *0..4 <\VBN\> ;
"are" *0..4 <\VBD\> ;
"''re" *0..4 <\VBN\> ;
"''re" *0..4 <\VBD\> ;
"am" *0..4 <\VBN\> ;
"am" *0..4 <\VBD\> ;
```

```
"''m" *0..4 <\VBN\> ;
"''m" *0..4 <\VBD\> ;
\PP\ *0..2 <\VBP\> ;
\NN\ *0..2 <\VBP\> ;
\NNS\ *0..2 <\VBP\> ;
\MD\ *0..1 <\VB\> ;
\MD\ \RB\ <\VB\> ;
```

## B.4   Spanish press corpus

```
\Z0,\ <\A\> [\ROA\ \ROP\];
\Z0,\ <\VOV\> [\ROA\ \ROP\];
"tal" * <\COS\>;
<\COS\> \IOV\;
<\COS\> \IOHV\;
<\COS\> \IOEV\;
"despus" [\ROA\ \ROP\] <\COS\>;
"antes" [\ROA\ \ROP\] <\COS\>;
\COS\ *0..8 [\COC\ \Z0,\] <\COS\>;
[\VOEV\ \UOEV\ \IOEV\ \GOEV\] \A\ <\COS\>;
[\VOV\ \UOV\ \IOV\ \GOV\] \A\ <\COS\>;
"ms" *1..3 <"que",\COS\>;
\Z0.\ \A\ <\COS\>;
"sino" <"que",\COS\>;
[\Z0\ \Z0\ \Z0!\ \ZOX\ \Z0,\ \Z0-\ \Z0.\ \Z0?\] <"sino",\COS\>;
[\Z0\ \Z0\ \Z0!\ \ZOX\ \Z0,\ \Z0-\ \Z0.\ \Z0?\] <"Sino",\COS\>;
"no" *0..9 <"sino",\COS\>;
"tanto" <"como", \COS\>;
[\VOEV\ \VOV\] *0..5 <"as",\D\>;
[\VOEV\ \VOV\] *0..5 <"as",\VOV\>;
<\D\> [\Z0\ \Z0\ \Z0!\ \ZOX\ \Z0,\ \Z0-\ \Z0.\ \Z0?\];
\VOV\ <\IOEV\>;
\N\ [\UOV\ \A\ \ROA\ \ROP\] <\POR\>;
\N\ [\TOD\ \TOI\ \TOO\ \TOP\ \TOQ\] <\POR\>;
\POR\ *0..8 "pero" <\POR\>;
"cosa" *0..2 <"que", \POR\>;
"Cosa" *0..2 <"que", \POR\>;
\POR\ *0..7 \COC\ <\POR\>;
"ms" <\N\> "que";
<"s",\VOV\> *0..1 [\TOI\ \POI\];
[\VOHV\ \UOHV\ \IOHV\ \GOHV\] \UOEV\ <\UOV\>;
```

# References

[Aarts & Korst 87] Aarts, E.H.L.; Korst, J.H.M.; *Boltzmann machines and their applications.* in de Bakker, J.W.; Nijman, A.J. and Treleaven, P.C. (eds). Proceedings PARLE (Parallel Architectures and Languages Europe). Lecture Notes in Computer Science Vol 258, 1987,pp.34-50

[Acebo et al. 94] Acebo, S.; Ageno, A.; Climent, S.; Farreres, J.; Padró, L.; Ribas, F.; Rodríguez, H.; Soler, O.; *MACO: Morphological Analyzer Corpus-Oriented* ESPRIT BRA-7315 Acquilex II, Working Paper 31, 1994

[Baum 72] Baum, L.E.; *An inequality and associated maximization technique in statistical estimation for probabilistic functions of a Markov process.* Inequalities, 3:1-8, 1972

[Brill 92] Brill, E.; *A simple rule-based part-of-speech tagger.* Proceedings ANLP 1992

[Briscoe et al. 94] Briscoe, E.J.; Greffenstette, G.; Padró, L.; Serail, I.; *Hybrid techniques for training Part-of-Speech taggers.* ESPRIT BRA-7315 Acquilex II, Working Paper 45, 1994

[Briscoe 94] Briscoe, E.J.; *Parsing (with) Punctuation and Shallow Syntactic Constraints on Part of Speech Sequences* Draft

[Bruce & Wiebe 94] Bruce, R.; Wiebe, J.; *A New Approach to Word Sense Disambiguation.* Proceedings ARPA 1994

[Cervell et al. 95] Cervell, S.; Climent, S.; Placer, R.; *Using MACO and MDS to tag a balanced corpus of Spanish* ESPRIT BRA-7315 Acquilex II, Working Paper, 1995

[Church & Hanks 90] Church, K.W.; Hanks, P. *Word association norms, mutual information and lexicography.* Computational Linguistics 16 (1) 22-29, 1990

[Cover & Thomas 91] Cover, T.M.; Thomas, J.A.; editors *Elements of information theory.* John Wiley 1991

[Cowie et al. 92] Cowie, J.; Guthrie, J.; Guthrie, L.; *Lexical Disambiguation using Simulated Annealing* Proceedings of DARPA Speech and Natural Language; Feb. 1992

[Cutting et al. 92] Cutting, D.; Kupiec, J.; Pederson, J.; Sibun, P.; *A Practical Part-of-Speech Tagger.* Proceedings ANLP 1992

[Elworthy 93] Elworthy, D.; *Part of Speech and Phrasal Tagging.* ESPRIT BRA-7315 Acquilex II, Working Paper 10, 1993

[Elworthy 94] Elworthy, D.; *Does Baum-Welch re-estimation help taggers?* Proceedings 4th ACL conference on Appliend Natural Language Processing, Stuttgart, 1994

[Gale et al. 92] Gale, W.; Church, K.W.; Yarowsky, D.; *Estimating Upper and Lowed Bounds on the Performance of Word Sense Disambiguation.* Proceedings ACL 1992

[Garside et al. 87] Garside, R.; Leech, G.; Sampson, G.; *The Computational Analysis of English.* Longman 1987

[Haralick 83] Haralick R.M.; *An interpretation for Probabilistic Relaxation.* Computer Vision, Graphics & Image Processing 22, pp. 388-395, 1983

[Harley 94] Harley, A.; *Cambridge language survey: Semantic tagger.* ESPRIT BRA-7315 Acquilex II, Working Paper 39, 1994

[Hirst 87] Hirst, G.; *Semantic Interpretation and the resolution of ambiguity* Cambridge University Press, 1987

[Hummel & Zucker 83] Hummel, R.A.; Zucker, S.W.; *On the foundations of relaxation labelling processes.* IEEE Transactions on Pattern Analysis and Machine Intelligence 5, n. 3 (1983)

[Kosko 90] Kosko, B.; *Neural Networks and Fuzzy Systems* Prentice-Hall 1990

[Lloyd 83] Lloyd, S.A.; *An optimization approach to relaxation labelling algorithms.* Image and Vision Computer, Vol.1, n.2, May 1983

[Miller et al. 91] Miller, G.A.; Beckwith, R.; Fellbaum, C.; Gross, D.; Miller, K.; *Five papers on WordNet.* International Journal of Lexicography, 1991

[Miller et al. 93] Miller, G.A.; Leacock, C.; Tengi, R.; Bunker, R.T.; *A semantic concordance* ARPA Workshop on Human Language Technology, 1993

[Miller et al. 94] Miller, G.A.; Chodorow, M.; Landes, S.; Thomas, R.G.; *Using a Semantic Concordance for Sense Identification.* Proceedings ARPA 1994 [Moreno-Torres 94]   Moreno-Torres, I.; *A morphological disambiguation tool (MDS). An application to Spanish.* ESPRIT BRA-7315 Acquilex II, Working Paper 24, 1994

[Pelillo & Refice 94]   Pelillo, M.; Refice M.; *Learning Compatibility Coefficients for Relaxation Labeling Processes.* IEEE Transactions on Pattern Analysis and Machine Intelligence 16, n. 9 (1994)

[Pelillo & Maffione 94]   Pelillo, M.; Maffione, A.; *Using Simulated Annealing to Train Relaxation Labelling Processes.* ICANN 1994

[Resnik 93]   Resnik, P.S.; *Selection and information: a class based approach to lexical relationships.* Ph.D. Thesis, Computer & Information Science Department, University of Pennsylvania.

[Ribas 94]   Ribas, F.; *An Experiment on Learning Appropiate Selectional Restrictions from a Parsed Corpora.* Proceedings of COLING 1994, Kyoto, Japan.

[Ribas 95]   Ribas, F.; *On Acquiring Appropiate Selectional Restrictions from Corpora Using a Semantic Taxonomy* Ph.D. Thesis. Dept. Llenguatges i Sistemes Informàtics, Universitat Politècnica de Catalunya, July 1995

[Richards et al. 81]   Richards, J.; Landgrebe, D.; Swain, P.; *On the accuracy of pixel relaxation labelling.* IEEE Transactions on System, Man and Cybernetics Vol. SMC-11, April 1981

[Rigau 94]   Rigau, G.; *An experiment on automatic semantic tagging of dictionary senses.* International Workshop on the future of the dictionary. Grenoble, France, 1994

[Rosenfeld et al. 76]   Rosenfeld, R.; Hummel, R.; Zucker, S.; *Scene labelling by relaxation operations.* IEEE Transactions on Systems, Man and Cybernetics. vol SMC 6, p420, 1976

[Schmid 94]   Schmid, H.; *Part of Speech Tagging with Neural Networks* COLING 94

[Southwell 40]   Southwell, R.; *Relaxation Methods in Engineering Science.* Clarendon, 1940

[Torras 89]   Torras, C.; *Relaxation and Neural Learning: Points of Convergence and Divergence.* Journal of Parallel and Distributed Computing 6, pp.217-244 (1989) [Viterbi 67]    Viterbi, A.J.; *Error bounds for convolutional codes and an asymptotically optimal decoding algorithm.* IEEE Transactions on Information Theory, pg 260-269, April 1967.

[Voutilainen 95]    Voutilainen, A.; *A syntax-based part-of-speech analyzer* Proceedings 7th EACL, 1995

[Waltz 75]    Waltz, D.; *Understanding line drawings of scenes with shadows.* Psycology of Computer Vision. P. Winston, New York: McGraw-Hill 1975

[Yarowsky 92]    Yarowsky, D.; *Word-sense disambiguations using statistical models of roget's categories trained on large corpora.* Proceedings of COLING 1992, Nantes, France.

[Yarowsky 93]    Yarowsky, D.; *One Sense per Collocation.* DARPA Workshop on Human Language Technology, Princeton, 1993

[Zucker et al. 78]    Zucker, S.W.; Krishnamurty, E.V.; Haar, R.L.; *Relaxation processes for scene labelling: Convergence, speed and stability.* IEEE Transactions on Systems, Man and Cybernetics 8, n. 1 (1978)

[Zucker et al. 81]    Zucker, S.W.; Leclerc, Y.G.; Mohammed, J.L.; *Continuous Relaxation and local maxima selection: Conditions for equivalence.* IEEE Transactions on Pattern Analysis and Machine Intelligence 3, n. 2 (1981)